\documentclass{article}
\usepackage{spconf,amsmath,graphicx,url}

\title{Recurrent Neural Network postfilters for Statistical Parametric speech synthesis}
\name{Prasanna Kumar Muthukumar, Alan W Black}
\address{Language Technologies Institute,\\ 
  Carnegie Mellon University, \\
  Pittsburgh, USA}

\begin{document}
%
\maketitle
\begin{abstract}
In the last two years, there have been numerous papers that have
looked into using Deep Neural Networks to replace the acoustic model
in traditional statistical parametric speech synthesis. However, far
less attention has been paid to approaches like DNN-based
postfiltering where DNNs work in conjunction with traditional acoustic
models. In this paper, we investigate the use of Recurrent Neural
Networks as a potential postfilter for synthesis. We explore the
possibility of replacing existing postfilters, as well as highlight the ease with
which arbitrary new features can be added as input to the
postfilter. We also tried a novel approach of jointly training the
Classification And Regression Tree and the postfilter, rather than the traditional approach of
training them independently. 
\end{abstract}
\section{Introduction}
\label{sec:intro}
Deep Neural Networks have had a tremendous influence on Automatic
Speech Recognition in the last few years. Statistical Parametric
Speech Synthesis\cite{zen2009statistical} has a tradition of borrowing
ideas from the speech recognition community\cite{dines2010measuring},
and so there has been a 
flurry of papers in the last two years on using deep neural networks
for speech
synthesis\cite{ze2013statistical,ling2013modeling,fan2014tts}. Despite this,
it would be difficult to argue as of now that deep neural networks have had the
same success in synthesis that they have had in ASR. DNN-influenced
improvements in synthesis have mostly been fairly moderate. This
becomes fairly evident when looking at the submissions to the Blizzard
Challenge\cite{Blizzard} in the last 3 years. Few of the submitted systems use deep
neural networks in any part of the pipeline, and those that do use
DNNs, do not seem to have any advantage over traditional well-trained
systems. 

Even in
cases where the improvements look promising, the techniques have had
to rely on the use of much larger datasets than is typically used. The
end result of this is that Statistical Parametric Synthesis ends up
having to lose the advantage it has over traditional unit-selection
systems\cite{hunt1996unit} in terms of the amount of data needed to
build a reasonable system. 

That being said, it is still be unwise to rule out the
possibility of DNNs playing an important role in speech synthesis
research in the future. DNNs are extremely powerful models, and like many
algorithms at the forefront of machine learning research, it might be
the case we haven't yet found the best possible way to use them. With
this in mind, in this paper we explore the
idea of using DNNs to \emph{supplement} existing systems rather than
as a replacement for any part of the system. More specifically, we will
explore the possibility of using DNNs as a {\bf postfilter} to try to
correct the errors made by the Classification And Regression Trees (CARTs). 

\section{Relation to prior work}

The idea of using a postfilter to fix flaws in the output of the
CARTs is not very new. Techniques such as Maximum Likelihood Parameter
Generation (MLPG)\cite{tokuda2000speech}
and Global variance\cite{tomoki2007speech} 
have become standard, and
even the newer ideas like the use of Modulation Spectrum\cite{takamichi2014postfilter} 
have started moving into the mainstream. These techniques provide
a significant improvement in quality, but suffer from the drawback
that the post-filter has to be derived 
\emph{analytically} for each feature that is used. MLPG for instance
exclusively deals with the means, the deltas, and the delta-deltas for
every state. Integrating non-linear combinations of the means across
frames or arbitrary new features like wavelets into MLPG will be
somewhat non-trivial, and requires 
revisiting the equations behind MLPG as well as a rewrite of the
code. 

Using a Deep Neural Network to perform this postfiltering can
overcome many of these issues. Neural networks are fairly agnostic to
the type of features provided as input. The input features can also  easily
be added or removed without having to do an extensive code rewrite. 

There has been prior work in using DNN based postfilters for
parametric synthesis in \cite{chen2014dnn} and
\cite{chen2015deep}. However, we differ from these in several
ways. One major difference is in the use of \emph{Recurrent} Neural
Networks (RNNs) as opposed to standard feedforward networks or
generative models like Deep Belief Nets. We believe that inherent
structure of RNNs is particularly suited to the time sensitive
nature of the speech signal. RNNs have been used before for synthesis
in \cite{fan2014tts}, but as a replacement for the existing acoustic
model and not as a postfilter.

In addition to this, we also explore the use of lexical features as
input to the postfilter. To our knowledge, this is the first attempt
at building a postfilter (neural network based or otherwise) that can
make use of text based features 
like phone and state information in addition to spectral features like
the MCEP means and standard deviations. We
also describe our efforts in making use of a novel algorithm called
Method of Auxiliary Coordinates (MAC) to {\bf jointly} train the
CARTs and the postfilter, rather than the traditional approach of
training the postfilter independent of the CART.

\section{Recurrent Neural Networks}

The standard feedforward neural network processes data one sample at a
time. While this may be perfectly appropriate for handling images,
data such as speech or video has an inherent time element that is
often ignored by these kind of networks. Each frame of a speech sample
is heavily influenced by the frames that were produced before
it. Both CARTs and feedforward networks generally tend to ignore these
inter-dependencies between consecutive frames. Crude approximations
like \emph{stacking} of frames are typically used to attempt to
overcome these problems. 

\begin{figure}[htb]

\begin{minipage}[b]{1.0\linewidth}
  \vspace{-0.2cm}
  \centering
  \centerline{\includegraphics[width=8.5cm]{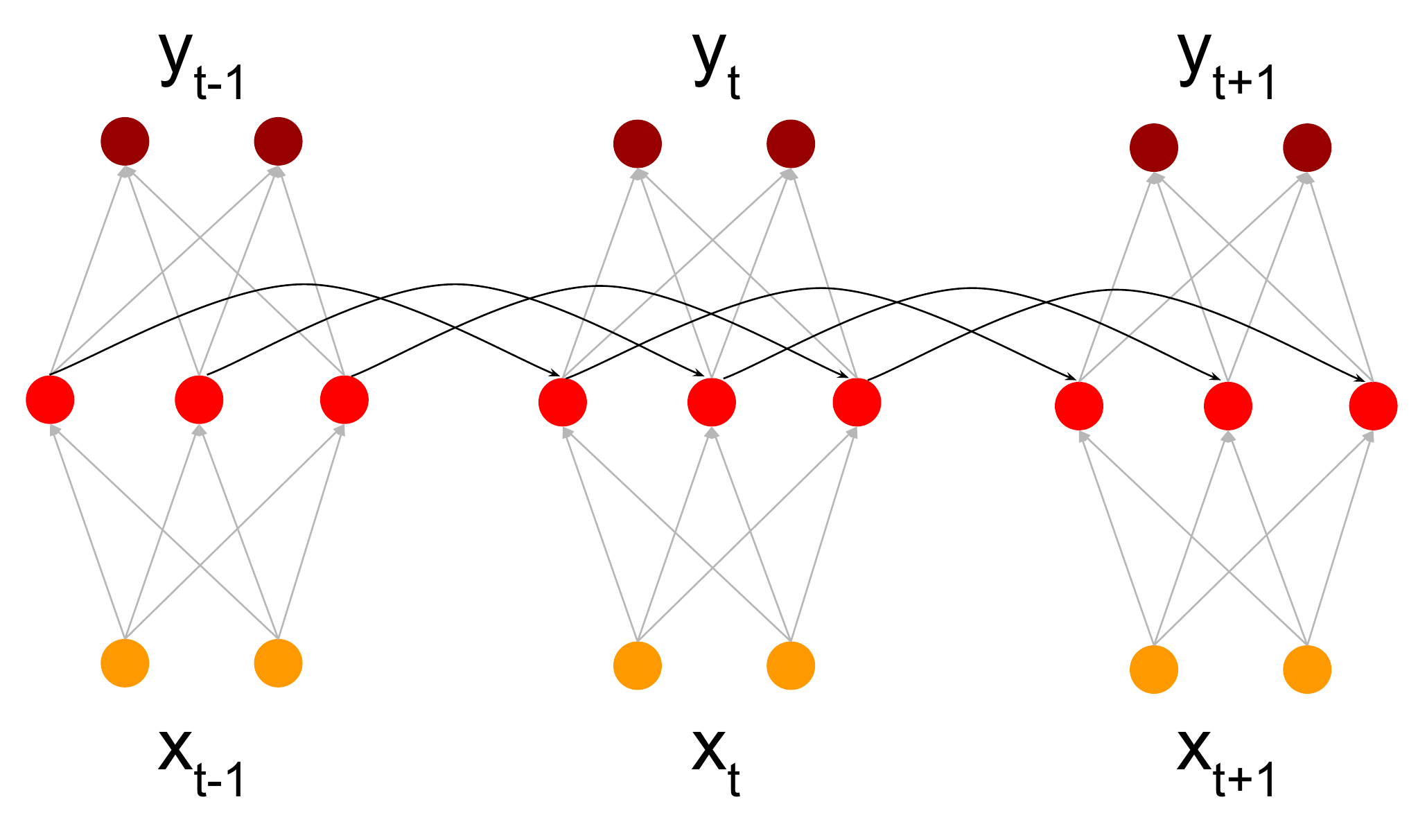}}
  \vspace{-1.0cm}
  \hspace{-1.0cm}
\end{minipage}
\caption{Basic RNN}
\label{fig:RNN}
\end{figure}

A more theoretically sound approach to handle the interdependencies
between consecutive frames is to use a \emph{Recurrent} Neural
Network\cite{elman1990finding}.  RNNs differ from basic feedforward
neural networks in their hidden layers. Each RNN hidden layer receives
inputs not only from its previous layer but also from activations of
itself for previous inputs. A simplified version of this structure is
shown in figure~\ref{fig:RNN}.  In the actual RNN that we used, every
node in the hidden layer is connected to the previous activation of
every node in that layer. However, most of these links have been omitted in the figure for
clarity. The structure we use for this paper is 
more or less identical to the one described in section 2.5 of
\cite{sutskever2013training}. 

To train an RNN to act as a postfilter, we start off by building a
traditional Statistical Parametric Speech Synthesis system. The
particular synthesizer we use is the Clustergen
system\cite{black2006clustergen}.  Once Clustergen has been trained on
a corpus in the standard way, we make Clustergen re-predict the entire
training data. As a result, we will now have Clustergen's predictions
of Mel Cepstral Coefficients (MCEPs) which is time aligned with the
original MCEPs extracted from speech. We then train an RNN to learn to
predict the original MCEP statics based on Clustergen's predictions of
the MCEP statics and deltas. This trained RNN can then be applied
to the CART's predictions of test data to remove some of the noise in
prediction. We use 25 dimensional MCEPs in our experiments. So the RNN takes 50 inputs
(25 MCEP statics + 25 deltas) and predicts 25 features (MCEP
statics). The results of doing this on four different corpora are 
shown in table~\ref{mcep_stat}.  Each voice had its own RNN. Mel Cepstral
Distortion\cite{kubichek1993mel} is used as the 
objective metric. 

\begin{table}[t,h]
\caption{\label{mcep_stat} {\it MCD without MLPG}}
\centerline{
\begin{tabular}{r c c } \hline
  Voice & Baseline & With RNN \\ \hline \hline
  RMS (1 hr)  & 4.98 & 4.91 \\ 
  SLT (1 hr) & 4.95 & 4.89 \\ 
  CXB (2 hrs) & 5.41 & 5.36 \\ 
  AXB (2 hrs) & 5.23 & 5.12 \\ \hline
\end{tabular}}
\end{table}

\begin{table}[h,t]
\vspace{-7mm}

\caption{\label{mcep_stat_mlpg} {\it MCD with MLPG}}
\centerline{
\begin{tabular}{ r c c } \hline
  Voice & Baseline & With RNN \\ \hline \hline
  RMS (1 hr)  & 4.75 & 4.79 \\ 
  SLT (1 hr)  & 4.70 & 4.75 \\ 
  CXB (2 hrs) & 5.16 & 5.23 \\ 
  AXB (2 hrs) & 4.98 & 5.01 \\ \hline
\end{tabular}}
\end{table}

RMS and SLT are voices from the CMU Arctic speech
databases\cite{kominek2004cmu}, about an hour of speech each. CXB is
an American female, and the 
corpus consists of various recordings from audiobooks. A 2-hour subset
of this corpus was used for the experiments described in this
paper. AXB is a 2-hour corpus of Hindi speech recorded by an Indian
female. RMS, SLT, and AXB were designed to be phonetically
balanced. CXB is not phonetically balanced, and is an audiobook unlike
the others which are corpora designed for building synthetic
voices.  

The RNN was implemented using the Torch7
toolkit\cite{collobert2011torch7}. The hyper-parameters were tuned on
the RMS voice for the 
experiment described in table~\ref{mcep_stat}. The result of this was
an RNN with 500 nodes in a single recurrent hidden layer, with
sigmoids as non-linearities, and a linear output
layer. To train the RNN, we used the ADAGRAD\cite{duchi2011adaptive}
algorithm along with a two-step BackPropagation Through
Time\cite{werbos1990backpropagation}. A batch size of 10, and a
learning rate of 0.01 were used. Early stopping was used for
regularization. L1, L2 regularization, and momentum did not improve
performance when used in addition to early stopping. No normalization
was done on either the output or the input MCEPs. The hyperparameters
were not tuned for any other voice, and  even the learning rate was
left unchanged. 

The results reported in table~\ref{mcep_stat} are with the MLPG option in
Clustergen turned off. The reason for this is that MLPG requires the
existence of standard deviations, in addition to the means of the
parameters that the CART predicts. There is no true set of standard
deviations that can be provided as training data for the RNN to learn
to predict; it only
has access to the true means in the training data. That being said, we
\emph{did} however apply MLPG by taking the MCEP  means from the
RNN, and standard deviations from the original CART predictions. We
found that it did not matter whether the means for the deltas were
predicted by the RNN or if they were taken from the CART predictions
themselves. The magnitudes of the deltas were typically so small that
it did not influence the RNN training as much as the statics did. So,
the deltas were omitted from the RNN predictions in favor of using the
deltas predicted by the CARTs directly. This also made the training a
lot faster as the size of the output layer was effectively halved. The
results of applying MLPG on the results from table~\ref{mcep_stat} are
shown in table~\ref{mcep_stat_mlpg}. Note that MLPG was applied on the
baseline system as well as the RNN postfilter system. 

\section{Adding lexical features}
In all of the previous experiments, the RNN was only using the same
input features that MLPG typically uses. This however does not
leverage the full power of the RNN. Any arbitrary feature can be fed as
input for the RNN to learn from. This is the advantage that an RNN has
over traditional postfiltering methods such as MLPG or Modulation
Spectrum. 

We added \emph{all} of Festival's lexical features as additional input
features for the RNN, in addition to CART's predictions of f0, MCEP
statics and deltas, and voicing. The standard deviations as well as
the means of the CART predictions were used. This resulted in 776
input features for the English language voices and 1076 features for
the Hindi one (the Hindi phoneset for synthesis is slightly
larger). The output features were the same 25-dimensional MCEPs from
the previous set of experiments. The results of applying this kind of RNN on
various corpora are shown in the following
tables. As for the previous set of experiments, each voice had its own RNN. Table~\ref{all_lex_no_MLPG} and table~\ref{all_lex_MLPG} show
results without and with MLPG respectively. In addition to the voices
tested in previous experiments, we also tested this approach on three additional
corpora, KSP (Indian male, 1 hour of Arctic), GKA (Indian male, 30mins
of Arctic), and AUP (Indian male, 30 mins of Arctic).

\begin{table}[t,h]
\caption{\label{all_lex_no_MLPG} {\it RNN with all lexical features. MLPG off}}
\centerline{
\begin{tabular}{ r c c } \hline
  Voice & Baseline & With RNN \\ \hline \hline
  RMS (1 hr)  & 4.98 & 4.77 \\ 
  SLT (1 hr) & 4.95 & 4.71 \\ 
  CXB (2 hrs) & 5.41 & 5.27 \\ 
  AXB (2 hrs) & 5.23 & 5.07 \\ 
  KSP (1 hr) & 5.13 & 4.88 \\ 
  GKA (30 mins) & 5.55 & 5.26  \\ 
  AUP (30 mins) & 5.37 & 5.09 \\ \hline
\end{tabular}}
\end{table}

\begin{table}[h,b,t]
\caption{\label{all_lex_MLPG} {\it RNN with all lexical features. MLPG on}}
\centerline{
\begin{tabular}{ r c c} \hline
  Voice & Baseline & With RNN \\ \hline \hline
  RMS (1 hr)  & 4.75 & 4.69 \\ 
  SLT (1 hr) & 4.70 & 4.64 \\ 
  CXB (2 hrs) & 5.16 & 5.15 \\ 
  AXB (2 hrs) & 4.98 & 4.98 \\ 
  KSP (1 hr) & 4.89 & 4.80 \\ 
  GKA (30 mins) & 5.27 & 5.17 \\ 
  AUP (30 mins) & 5.10 & 5.02 \\ \hline
\end{tabular}}
\end{table}

As can be seen in the tables, the RNN \emph{always} gives an
improvement when MLPG is turned off. \cite{kominek2008synthesizer}
reports that an MCD decrease of 0.12 is equivalent to doubling the
amount of training data. The MCD decrease in
table~\ref{all_lex_no_MLPG} is far beyond that in all of the voices
that are tested. It is especially heartening to see that the MCD
decreases significantly even in the case where the corpus is only 30
minutes. 
With MLPG turned on, the RNN always improves the system or is no
worse. The decrease in MCD is much smaller though. 

With MLPG turned on, the decrease in MCD with the RNN system is not large enough for
humans to be able to do reasonable listening tests. With MLPG turned
off though, the RNN system was shown to be vastly better in informal
listening tests. But our goal was to
build a system much better than the baseline system \emph{with}
MLPG, and so no formal listening tests were done.

\section{Joint training of the CART and the postfilter}

The traditional way to build any postfilter for Statistical Parametric
Speech Synthesis is to start off by building the Classification And
Regression Trees that predict
the parameters of the speech signal, and then train or apply the
postfilter on the output of the CART. The drawback of this is that the CART is
unnecessarily agnostic to the existence of the postfilter. CARTs and
postfilters need not necessarily work well together, given that each
has its own set of idiosyncracies when dealing with data. MCEPs might
not be the best representation that connects these two. In an ideal system, the CART and
the postfilter should jointly agree upon a representation of the
data. This representation should be easy for the CART to learn, as
well as reasonable for the postfilter to correct. 

One way of achieving this goal is to use a fairly new technique called
Method of Auxiliary Coordinates
(MAC)\cite{carreira2012distributed}. To understand how this technique
works, we need to mathematically formalize our problem. 

\begin{figure}[htb]
  \centering
  \centerline{\includegraphics[width=\linewidth]{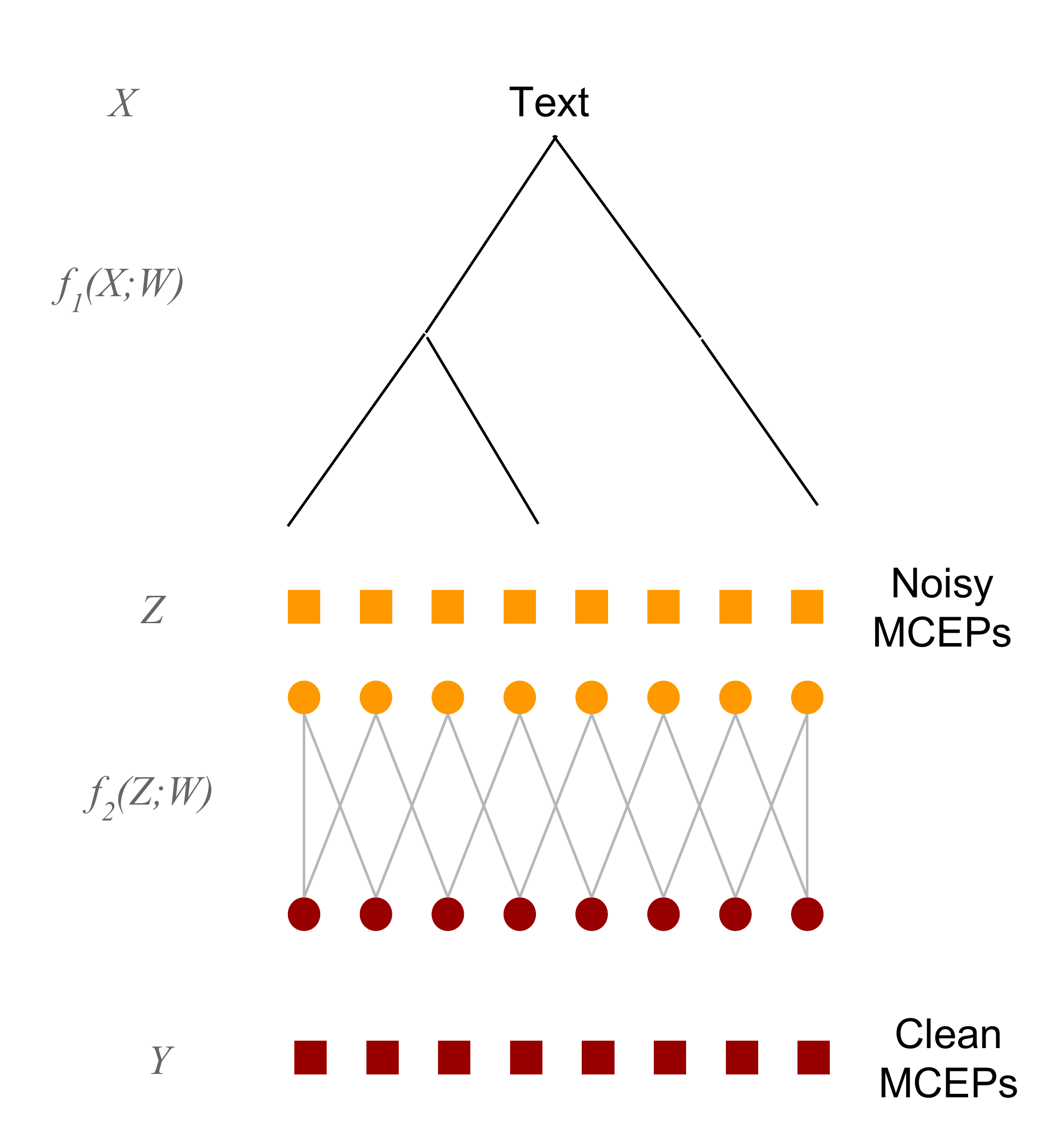}}
  \caption{{\it Joint training of the CART and the RNN}}
  \label{mac}
\end{figure}

Let $X$ represent the linguistic features that are input to the
CART. Let $Y$ be the correct MCEPs that need to be predicted at the
end of synthesis. Let $f_{1}()$ be the function that the CART represents, and $f_{2}()$ the function that the RNN represents. The CART and the RNN both
have parameters that are learned 
as part of training. These can be concatenated into one set of
parameters $W$. The objective
function we will want to minimize can therefore be written as:
\begin{eqnarray*}
E(W) = \| f_{2}(f_{1}(X;W);W) - Y \|^{2}
\end{eqnarray*}
The MAC algorithm basically rewrites the above equation to make it
easier to solve. This is done by explicitly defining an intermediate
representation $Z$ that acts as the target for the CARTs and as input
to the RNN. The previous equation can now be rewritten as:
\begin{eqnarray*}
E(W,Z) = \| f_{2}(Z;W) - Y\|^{2} \\
s.t.\; Z = f_{1}(X;W)
\end{eqnarray*}

The hard constraint in the above equation is then converted to a
quadratic penalty term:
\begin{eqnarray*}
E(W,Z;\mu) = \| f_{2}(Z;W) - Y\|^{2} + \mu \|f_{1}(X;W) - Z \|^{2}
\end{eqnarray*}
where $\mu$ is slowly increased towards $\infty$.

The original objective function in the first equation only had the
parameters of the CART and the RNN as variables to be used for the
minimization. The rewritten objective function adds the intermediate
representation $Z$ as an auxiliary variable that will also be used in
the minimization. We minimize the objective function by alternatingly
optimizing the parameters $W$ and the intermediate variables
$Z$. Minimizing with respect to $W$ is more or less equivalent to
training the CARTs and RNNs independently using a standard RMSE
criterion. Minimization with respect to $Z$ is done through Stochastic
Gradient Descent. Intuitively, this alternating minimization has the
effect of alternating between 
optimizing the CART and RNN, and optimizing for an intermediate
representation that works well with both. 

We applied this algorithm to the RNN and CARTs built for the SLT
 and RMS voices built for table~\ref{mcep_stat}. The results of this are
shown in table~\ref{MAC_results}. Starting from the second iteration, any
further optimization of either the $W$ or the $Z$ variables only results in a
very small decrease in the value of the objective function. So, we did
not run the code past the second iteration. 

We did not try MAC for the RNNs which use lexical features. This is
because running the $Z$ optimization on those RNNs would have given us
a new set of lexical features for which we would have no way of
extracting from text.

\begin{table}[h,b,t]
\caption{\label{MAC_results} {\it MCD Results of MAC}}
\centerline{
\begin{tabular}{ c c c} \hline
  Method & SLT & RMS  \\ \hline \hline
  RNN from table 1   & 4.89 & 4.91\\
  1st MAC iteration  & 4.86 & 4.92 \\
  2nd MAC iteration  & 4.89 & 4.92 \\ \hline
\end{tabular}}
\end{table}

On our experiments on the RMS and SLT voices, there was no
significant improvement in MCD. There are marginal MCD
improvements in the 1st iteration for the SLT voice but these are not
significant. Preliminary experiments 
suggest that one reason for the suboptimal performance is the
overfitting towards the training 
data. It is difficult to say why the MAC algorithm does not perform very well
in this framework. The search space for the parameters and the number
of possible experiments that can be run is extremely large though, and
so it is likely that a more thorough investigation will provide
positive results. 

\section{Discussion}
In this paper, we have looked at the effect of using RNN based
postfilters. The massive improvements we get in the absence of other
postfilters such as MLPG indicate that the RNNs are definitely a
viable option in the future, especially because of the ease with which
random new features can be added. However the combination of MLPG and RNNs
are slightly less convincing. This could mean that the RNNs have
learned to do approximately the same thing as MLPG does. Or it could
mean that MLPG is not really appropriate to be used on RNN outputs. We
believe that the answer might actually be a combination of both. The right
solution might ultimately be to find an algorithm akin to MAC which
can tie various postfilters together for joint training. Future
investigations in these directions might lead to insightful new
results.


\section{Acknowledgements}
Nearly all experiments reported in this paper were run on a Tesla K40
provided by an Nvidia hardware grant, or on g2.8xlarge EC2 instances
funded by an Amazon AWS research grant. In the absence of these, it would have
been impossible to run all of the experiments described above in a
reasonable amount of time. 
\vfill\pagebreak

\bibliographystyle{IEEEbib}
\bibliography{refs}

\begin{thebibliography}{10}

\bibitem{zen2009statistical}
Heiga Zen, Keiichi Tokuda, and Alan~W Black,
\newblock ``Statistical parametric speech synthesis,''
\newblock {\em Speech Communication}, vol. 51, no. 11, pp. 1039--1064, 2009.

\bibitem{dines2010measuring}
John Dines, Junichi Yamagishi, and Simon King,
\newblock ``Measuring the gap between {HMM}-based {ASR} and {TTS},''
\newblock {\em Selected Topics in Signal Processing, IEEE Journal of}, vol. 4,
  no. 6, pp. 1046--1058, 2010.

\bibitem{ze2013statistical}
Heiga Zen, Andrew Senior, and Mike Schuster,
\newblock ``Statistical parametric speech synthesis using deep neural
  networks,''
\newblock in {\em Acoustics, Speech and Signal Processing (ICASSP), 2013 IEEE
  International Conference on}. IEEE, 2013, pp. 7962--7966.

\bibitem{ling2013modeling}
Zhen-Hua Ling, Li~Deng, and Dong Yu,
\newblock ``Modeling spectral envelopes using restricted boltzmann machines and
  deep belief networks for statistical parametric speech synthesis,''
\newblock {\em Audio, Speech, and Language Processing, IEEE Transactions on},
  vol. 21, no. 10, pp. 2129--2139, 2013.

\bibitem{fan2014tts}
Yuchen Fan, Yao Qian, Fenglong Xie, and Frank~K Soong,
\newblock ``{TTS} synthesis with bidirectional {LSTM} based recurrent neural
  networks,''
\newblock in {\em Proc. Interspeech}, 2014, pp. 1964--1968.

\bibitem{Blizzard}
``{B}lizzard {C}hallenge 2015,''
  \url{http://www.synsig.org/index.php/Blizzard_Challenge_2015}.

\bibitem{hunt1996unit}
Andrew~J Hunt and Alan~W Black,
\newblock ``Unit selection in a concatenative speech synthesis system using a
  large speech database,''
\newblock in {\em Acoustics, Speech, and Signal Processing, 1996. ICASSP-96.
  Conference Proceedings., 1996 IEEE International Conference on}. IEEE, 1996,
  vol.~1, pp. 373--376.

\bibitem{tokuda2000speech}
Keiichi Tokuda, Takayoshi Yoshimura, Takashi Masuko, Takao Kobayashi, and
  Tadashi Kitamura,
\newblock ``Speech parameter generation algorithms for {HMM}-based speech
  synthesis,''
\newblock in {\em Acoustics, Speech, and Signal Processing, 2000. ICASSP'00.
  Proceedings. 2000 IEEE International Conference on}. IEEE, 2000, vol.~3, pp.
  1315--1318.

\bibitem{tomoki2007speech}
Tomoki Toda and Keiichi Tokuda,
\newblock ``A speech parameter generation algorithm considering global variance
  for {HMM}-based speech synthesis,''
\newblock {\em IEICE TRANSACTIONS on Information and Systems}, vol. 90, no. 5,
  pp. 816--824, 2007.

\bibitem{takamichi2014postfilter}
Shinnosuke Takamichi, Tomoki Toda, Graham Neubig, Sakriani Sakti, and Shigenari
  Nakamura,
\newblock ``A postfilter to modify the modulation spectrum in {HMM}-based
  speech synthesis,''
\newblock in {\em Acoustics, Speech and Signal Processing (ICASSP), 2014 IEEE
  International Conference on}. IEEE, 2014, pp. 290--294.

\bibitem{chen2014dnn}
Ling-Hui Chen, Tuomo Raitio, Cassia Valentini-Botinhao, Junichi Yamagishi, and
  Zhen-Hua Ling,
\newblock ``{DNN}-based stochastic postfilter for {HMM}-based speech
  synthesis,''
\newblock {\em Proc. Interspeech, Singapore, Singapore}, 2014.

\bibitem{chen2015deep}
Ling-Hui Chen, Tuomo Raitio, Cassia Valentini-Botinhao, Zhen-Hua Ling, and
  Junichi Yamagishi,
\newblock ``A deep generative architecture for postfiltering in statistical
  parametric speech synthesis,''
\newblock {\em Audio, Speech, and Language Processing, IEEE/ACM Transactions
  on}, vol. 23, no. 11, pp. 2003--2014, 2015.

\bibitem{elman1990finding}
Jeffrey~L Elman,
\newblock ``Finding structure in time,''
\newblock {\em Cognitive science}, vol. 14, no. 2, pp. 179--211, 1990.

\bibitem{sutskever2013training}
Ilya Sutskever,
\newblock {\em Training recurrent neural networks},
\newblock Ph.D. thesis, University of Toronto, 2013.

\bibitem{black2006clustergen}
Alan~W Black,
\newblock ``{CLUSTERGEN}: a statistical parametric synthesizer using trajectory
  modeling.,''
\newblock in {\em INTERSPEECH}, 2006.

\bibitem{kubichek1993mel}
Robert~F Kubichek,
\newblock ``Mel-cepstral distance measure for objective speech quality
  assessment,''
\newblock in {\em Communications, Computers and Signal Processing, 1993., IEEE
  Pacific Rim Conference on}. IEEE, 1993, vol.~1, pp. 125--128.

\bibitem{kominek2004cmu}
John Kominek and Alan~W Black,
\newblock ``The {CMU} {A}rctic speech databases,''
\newblock in {\em Fifth ISCA Workshop on Speech Synthesis}, 2004.

\bibitem{collobert2011torch7}
Ronan Collobert, Koray Kavukcuoglu, and Cl{\'e}ment Farabet,
\newblock ``Torch7: A matlab-like environment for machine learning,''
\newblock in {\em BigLearn, NIPS Workshop}, 2011, number EPFL-CONF-192376.

\bibitem{duchi2011adaptive}
John Duchi, Elad Hazan, and Yoram Singer,
\newblock ``Adaptive subgradient methods for online learning and stochastic
  optimization,''
\newblock {\em The Journal of Machine Learning Research}, vol. 12, pp.
  2121--2159, 2011.

\bibitem{werbos1990backpropagation}
Paul~J Werbos,
\newblock ``Backpropagation through time: what it does and how to do it,''
\newblock {\em Proceedings of the IEEE}, vol. 78, no. 10, pp. 1550--1560, 1990.

\bibitem{kominek2008synthesizer}
John Kominek, Tanja Schultz, and Alan~W Black,
\newblock ``Synthesizer voice quality of new languages calibrated with mean mel
  cepstral distortion,''
\newblock in {\em Spoken Languages Technologies for Under-Resourced Languages},
  2008.

\bibitem{carreira2012distributed}
Miguel~A Carreira-Perpin{\'a}n and Weiran Wang,
\newblock ``Distributed optimization of deeply nested systems,''
\newblock {\em arXiv preprint arXiv:1212.5921}, 2012.

\end{thebibliography}

\end{document}